\documentclass[
]{ceurart}

\usepackage{listings}
\usepackage{enumitem}

\begin{document}

\copyrightyear{2024}
\copyrightclause{Copyright for this paper by its authors.
  Use permitted under Creative Commons License Attribution 4.0
  International (CC BY-NC-ND 4.0).}

\conference{\textbf{This paper was presented in IFOW, Joint Ontology Workshops (JOWO) at University of Twente, July 2024} \\
\\
Formal Ontology in Information Systems Conference (FOIS) '24: Integrated Food Ontology Workshop (IFOW)\\
July 15--19, 2024, University of Twente, Enschede, Netherlands} 

\vspace*{-2.0 cm}
\title{Building FKG.in: a Knowledge Graph for Indian Food }


\author[1]{Saransh Kumar Gupta}[%
orcid=0009-0000-5887-2301,
email=saransh.gupta@ashoka.edu.in,
url=,
]
\cormark[1]
\fnmark[1]
\address[1]{Ashoka University, India}

\author[1]{Lipika Dey}[%
orcid=0000-0003-3831-5545,
email=lipika.dey@ashoka.edu.in,
url=,
]
\cormark[1]
\fnmark[1]

\author[1]{Partha Pratim Das}[%
orcid=0000-0003-1435-6051,
email=partha.das@ashoka.edu.in,
url=,
]
\cormark[1]
\fnmark[1]

\author[2]{Ramesh Jain}[%
orcid=0000-0003-2373-4966,
email=jain49@gmail.com,
url=,
]
\fnmark[1]
\address[2]{Institute of Future Health, UC Irvine, USA}

\cortext[1]{Corresponding author.}
\fntext[1]{These authors contributed equally.}

\begin{abstract}
  This paper presents an ontology design along with knowledge engineering, and multilingual semantic reasoning techniques to build an automated system for assimilating culinary information for Indian food in the form of a knowledge graph. The main focus is on designing intelligent methods to derive ontology designs and capture all-encompassing knowledge about food, recipes, ingredients, cooking characteristics, and most importantly, nutrition, at scale. We present our ongoing work in this workshop paper, describe in some detail the relevant challenges in curating knowledge of Indian food, and propose our high-level ontology design. We also present a novel workflow that uses AI, LLM, and language technology to curate information from recipe blog sites in the public domain to build knowledge graphs for Indian food. The methods for knowledge curation proposed in this paper are generic and can be replicated for any domain. The design is application-agnostic and can be used for AI-driven smart analysis, building recommendation systems for Personalized Digital Health, and complementing the knowledge graph for Indian food with contextual information such as user information, food biochemistry, geographic information, agricultural information, etc.
\end{abstract}


\maketitle
\vspace*{-0.5cm}
\section{Introduction}
\vspace*{-0.2cm}
Food is playing an increasingly central role in health and sustainability discourses as the preservation of diverse cultures, food security, precision nutrition, personal and public health, agricultural practices, climate impact, and supply chains become focal points of discussion. However, any definitive effort in such a scientific pursuit requires well-founded applications to be designed around food and food-related data, with access to knowledge representation and reasoning systems for food. In this light, food knowledge graphs are crucial and reusable digital resources that can capture various nuances of food including but not limited to recipes, ingredients, flavor, texture, cooking techniques, cuisine, nutritional information, and mealtimes. They can be used for various applications like food recommendation, recipe recommendation, diet planning, health tracking, food quality control, managing food supply chains, and so on. While several countries including the US, parts of Europe (like Latvia, Norway, Spain, the UK, Italy, and Portugal), China, and Japan are working on building such knowledge bases for specific regions, there appears to be a vacuum when it comes to Indian food. We intend to bridge this gap to aid food computing initiatives for Indian food.

In this paper, we present our work on building a knowledge graph for Indian food, named FKG.in, which aims to exhaustively cover the panorama of Indian food and act as a digital resource for building subsequent food computing applications over it. The proposed ontology adapts from earlier food ontologies along with modifications and extensions to capture unique aspects of Indian food and is designed in an application-agnostic way. We also propose a novel AI-based semi-automated approach to curate culinary information from multiple websites in the public domain to populate the knowledge graph, along with a human-in-the-loop intervention to ensure the soundness of information.

The rest of the paper is organized as follows. Section 2 presents an overview of the work done in the areas of designing food ontology and building food knowledge graphs, along with earlier efforts in building Indian food knowledge graphs. Section 3 presents the unique challenges associated with the task of consolidating knowledge about Indian food. Section 4 presents the ontology design and its connections to existing food knowledge graphs. Section 5 presents the AI-based technologies adopted to populate a knowledge base for Indian food. Section 6 presents some results. Finally, we conclude with plans to extend the knowledge graph and integrate it with other food computing applications.

\vspace*{-0.2cm}
\section{Related work}
\vspace*{-0.2cm}
Several food knowledge graphs have been constructed based on these ontologies and their extensions, to support food computing applications. We mention only a few representative ones here. For a more comprehensive review of food ontology, food knowledge graphs, and food computing applications, one may refer to the works cited in \cite{Min19,Min20,Min22}. In \cite{Min22}, authors have categorized existing food knowledge graphs into four different types - (1) knowledge graphs about recipes, (2) knowledge graphs about nutrients and health, (3) knowledge graphs about food safety, and (4) general food knowledge graphs. We follow the same pattern to group both ontology and knowledge graphs, though there exist many overlaps among the groups.
  
The first group of ontology mostly focuses on concepts of food related to recipes and cooking. Notable in this group are Table \cite{Cordier14}, Cooking ontology \cite{Ribeiro06}, BBC food ontology\footnote{https://www.bbc.co.uk/ontologies/food-ontology}, and so on. There are a few ontologies dedicated to special categories of food, like Open Food Facts\footnote{https://world.openfoodfacts.org/data} that model information about packaged foods, Seafood ontology \cite{Sherimon21}, and so on.  

Recipe knowledge graphs are built to store recipe entities that are extracted from crowdsourced consumer review sites, recipe-sharing websites, and social media to primarily support food recommendation systems and build social networks around food. Foodbar knowledge graph \cite{Zulaika18} is one such system that extracts consumer opinions ratings etc. from different sources and augments this information with information about users, points of interest, cultural facts, and so on. RcpKG \cite{Lei21} is a multimodal and hierarchical food knowledge graph that curates information from popular recipe websites like Yummly and AllRecipes as well as semi-structured datasets like Recipe1M+ \cite{Marin21}. RcpKG also incorporates social relationships into the food knowledge graph for generating food recommendations that can take care of both personal preferences and social relationships. In a unique experiment, cooking is viewed as a uniquely human endeavor for transforming raw ingredients into delicious dishes \cite{Bagler20}, and it is proposed that recipes can be viewed as cultural capsules that capture culinary protocols. This work is focused on learning the valid protocols for a given set of constraints and thereby generating recipes without violating the cooking principles. This work is also envisaged to generate recipes following the culinary grammar that can be leveraged to improve public health through dietary interventions. The underlying knowledge base is RecipeDB\footnote{https://cosylab.iiitd.edu.in/recipedb/} \cite{Batra20}, which is a structured compilation of recipes, and ingredients along with their nutrition and flavor profiles and health associations. 

Several ontologies have been built to cater to the concepts of food, nutrition, and health. Personalized Information Platform for Health and Life Services (PIPS) \cite{Cantais05} consists of an abstract model of different types of food along with health and nutrition concepts, targeted at providing nutritional advice for diabetic patients. FOODS \cite{Snae08} also focused on storing information about food and nutrition with an eye toward food or menu planning for people with diabetes. Edamam food ontology\footnote{https://www.edamam.com/} provides concepts related to food, recipes, and nutrition to promote healthy eating through various applications like cooking robots. FoodOn \cite{Dooley18} contains a fairly exhaustive list of properties that can be used to describe basic ingredient types coming from animal, plant, or fungal origins, agricultural and animal husbandry practices linked to their growth, also lists of common processed food items, and chemical ingredients along with the processes used to make them and terminology to describe nutritional values. The ontology aims to provide a shared vocabulary that can be used for knowledge exchange across domains like environment, agriculture, animal husbandry, food processing, etc. to ensure food safety and security. 

FoodKG \cite{Haussmann19} is a large-scale and unified food knowledge graph that brings together FoodOn and WhatToMake ontology and contains recipe and nutrient instances extracted from Recipe1M+ as well as nutrient records from the US Department of Agriculture (USDA). This knowledge graph can support a multitude of applications, like recipe recommendations, ingredient substitutions, and Question Answering about nutrition. The Chinese Food Knowledge Graph \cite{Chi18} containing information about Chinese dietary cultural elements and Traditional Chinese Medicine was built to enable knowledge retrieval about health and balanced diets. 

Other special-purpose ontologies include AGROVOC \cite{Caracciolo11}, which is dedicated to storing agriculture, fisheries, and forestry terminologies related to food. Food Track and Trace Ontology \cite{Pizzuti14}, which models knowledge related to the food supply chain, has been designed specially for the food safety domain to help in food traceability. Supply Chain Traceability (SCT) ontology \cite{Ameri22} also supports information about critical tracking events (CTEs) to provide unified support to food traceability from logistics to production lines. The Meat Supply Chain Ontology (MESCO) is specialized to support the meat supply chain area. Knowledge graphs built along these lines include the Food safety knowledge graph \cite{Qin19} and the Food spot-check knowledge graph \cite{Qin20} which are mainly concerned with food safety issues. The Food Safety Knowledge Graph supports a Question Answering application to answer user queries about unqualified foods, based on official information released on the Internet. Food spot-check supports a similar application based on data released on the Internet about spot-checks. 

In the Indian context, a framework for knowledge acquisition, conceptualization, formalization, implementation, and evaluation for a knowledge base is presented in \cite{Padmavathi16}. This knowledge base contained many Indian food items, but the focus was on methodology. A digital resource of 528 key Indian food ingredients along with their nutritional information is curated and presented in \cite{Sahu22}. 
Most of the above ontologies and knowledge graphs were built from semi-structured recipe cards for specific applications. \cite{Madalli17} proposed a formal but generic methodology for gathering information and building a food ontology in an automated fashion. The proposed work for building an Indian Food Ontology extends this pipeline.

\vspace*{-0.2cm}
\section{Unique and Complex Challenges in building FKG.in: Significance and Relevance of Indian food}
\vspace*{-0.2cm}
The diverse Indian cuisine reflects a history of over 8,000 years, during which the history of various ethnic groups and cultures have interacted with each other in the Indian subcontinent, resulting in a vast variety of cooking techniques, flavors, and regional cuisines. While this diversity has resulted in a rich repertoire of recipes, it has also introduced some unique challenges towards automating the task of building a food knowledge base. We note these and a few other challenges below:
\vspace*{-0.1cm}
\begin{enumerate}[noitemsep]
    \item \textbf{Lack of a comprehensive vocabulary} of food items has necessitated that this work start from almost scratch. This spans all concepts related to food like ingredients, cuisines, styles, cooking processes, cookware, and so on. There exists \textbf{a multiplicity of recipes} with the same name but different compositions from different regions. To a large extent, this occurs due to regional variations in climate, culture, and availability of ingredients. The most common example of this is \textit{dal} (lentil soup) which, though an integral part of almost all regional meals, has huge variations across the country. Additionally, as Indian food recipes are often quite complex, capturing the nuances of similar recipes is often a very difficult task.

    \item The \textbf{multilingual} nature of India poses a challenge exactly opposite to the previous one. The same food items have various vernacular names across the country. For example, \textit{haldi} (Hindi), \textit{holud} (Bengali), \textit{halad} (Marathi), \textit{pasupu} (Telugu) and \textit{manjal} (Tamil), all refer to turmeric in different Indian languages. Building a common and inclusive dictionary of food items for India needs multilingual capabilities to address this diversity.
    

    \item \textbf{Food homonyms} present a challenge in the form of confusing granularities where ingredients and recipes may be known by the same name. A typical example is \textit{chawal} (rice) which refers to both raw rice i.e. an ingredient and steamed rice i.e. the final dish.

    \item Another challenge stems from the fact that Indian food is \textbf{not about precision cooking}. Measurements are often expressed in terms of common kitchen containers like “a cup” or “a \textit{katori} (bowl)”, for which there are no specific standards. Use of linguistic variables like “a little”, “some”, and “a handful of” are also encountered quite often.

    \item \textbf{Socio-cultural association of food items with festivals, religious celebrations, and spiritual motivations} is a worldwide phenomenon. These notions have to be captured to generate contextually relevant recommendations. For example, \textit{kheer} or \textit{payasam} (milk pudding), and \textit{Hyderabadi haleem} are almost always associated with different religious celebrations, whereas abstinence food is typically expected to be without garlic or onion. 
    
    \item \textbf{Temporal association of a dish} are derived from Indian food habits, which are largely society-driven or family-driven, which has led to unique styles and practices related to \textit{what is usually eaten when}, that are crucial to Indian food, but not properly documented.
    \item \textbf{Absence of any comprehensive source of nutritional information} about Indian food makes it difficult to envision well-grounded applications in health around Indian food.  
\end{enumerate}

In this paper, we have proposed an AI-driven pipeline to address some of these challenges. 

\vspace*{-0.2cm}
\section{Proposed Ontology Design for Indian Food}
\vspace*{-0.2cm}

We now present the details of FKG.in, the Indian Food Ontology, which is inspired by FoodOn \cite{Dooley18} and FoodKG \cite{Haussmann19}, and wherever needed, adapted them to suit the Indian context. FKG.in attempts to capture important properties of Indian food in terms of culinary language, cooking variations, and precision nutrition. In doing so, we have also attempted to make the ontology modular and flexible to incorporate changes in the knowledge curation stage if required.

\begin{figure}
  \centering
  \includegraphics[width=\linewidth]{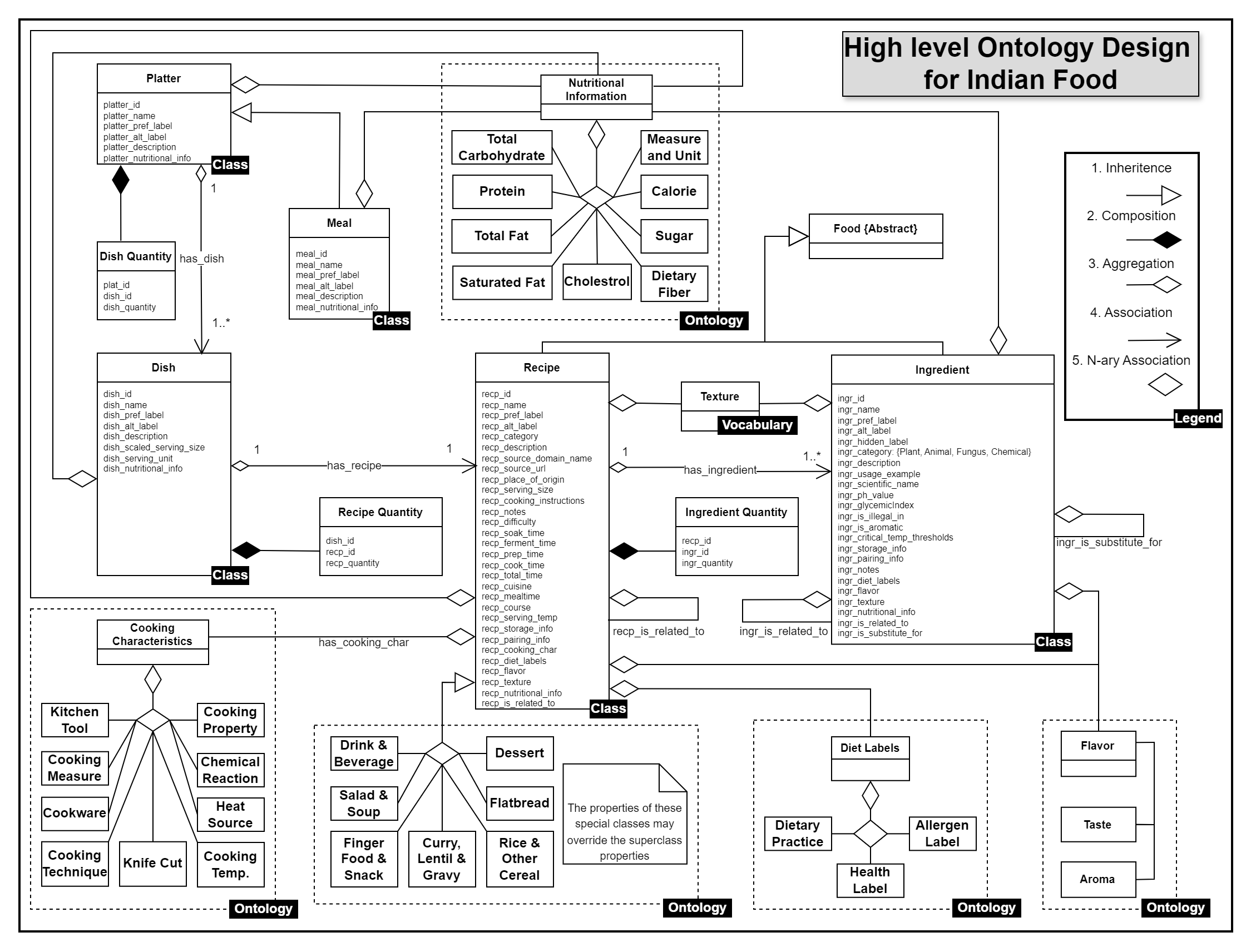}
  \caption{High-level Ontology Design for Indian Food (incl. Cooking Characteristics and Nutrition)}
  \label{fig:ontology_design_for_indian_food}
\end{figure}

Figure \ref{fig:ontology_design_for_indian_food} presents the proposed ontology design. \textbf{Food} is an abstract superclass and \textbf{Ingredients} and \textbf{Recipes} are its most important subclasses. These two concepts are described in more detail below:
\vspace*{-0.1cm}
\begin{enumerate}[noitemsep]
    \item \textbf{Recipe} class: At the heart of the proposed food ontology lies the \textbf{Recipe} class. A \textbf{Recipe} instance (or recipe) is composed of measured ingredients, physical and conceptual properties, cooking characteristics, and a set of cooking instructions. Different sets of properties are associated with the \textbf{Recipe} class. The first set comprises those that have values represented as simple strings like \textbf{name}, \textbf{cuisine}, \textbf{serving size}, \textbf{calories}, etc. Secondly, a recipe is characterized by detailed \textbf{cooking instructions}, which we store as long text. The third set of properties is composite. For example, a recipe has \textbf{Cooking Characteristics}, which is an aggregate class, designed along the lines of a similar class available in FoodKG \cite{Haussmann19}. This contains \textbf{cooking techniques}, \textbf{cookware}, \textbf{cooking temperature}, etc. Vocabularies to cater to Indian cooking processes, practices, kitchen utensils, etc. have been curated. For a \textbf{Recipe} instance, while many of these property values may be available from data, some are derived using predefined functions. The most important properties in this set are \textit{cuisine}, \textit{nutritional information} \textit{diet labels} and so on. These properties are derived from the ingredients, their measurements, and cooking characteristics using dedicated functions. The property \textbf{pairing information} stores names of other recipes that are usually taken with it. For example, <\textit{chawal}, \textit{dal}>, <\textit{idli}, (\textit{chutney}, \textit{sambar})>, <\textit{biryani}, \textit{raita}/\textit{salan}> are common pairs in Indian food.

    \item \textbf{Recipe} subclasses: There are several subclasses of \textbf{Recipe} like \textbf{Flatbread}, \textbf{Dessert}, \textbf{Beverage \& Drink}, etc. which have typical properties associated with them like \textbf{texture}, \textbf{serving temperature}, and so on. This set of subclasses has been adapted from FoodKG \cite{Haussmann19} and extended to accommodate Indian food items like \textit{curry, bharta} etc.

    \item \textbf{Ingredient} class: A \textbf{Recipe} instance uses \textbf{ingredients}, which is the second most significant concept in our ontology. Any food item that can contribute to a combination of other ingredients to make a particular recipe is an instance of the \textbf{Ingredient} class. This class has a long list of properties that define the \textbf{origin}, \textbf{flavor}, \textbf{glycemic index}, \textbf{nutritional information}, \textbf{pH value}, etc. This list of properties is also adapted from the class \textbf{Food Category} of FoodKG \cite{Haussmann19} and extended to accommodate Indian food concepts. According to \cite{Haussmann19}, there are four basic ingredient categories based on origin - \textbf{plant}, \textbf{animal}, \textbf{fungus}, and \textbf{chemical}. Plant-origin ingredients themselves may be used either in primary or processed forms. Further sub-categorization may be based on \textbf{fruits}, \textbf{herbs}, \textbf{legumes}, \textbf{milled cereal products}, and so on. There are many more sub-categories highlighting texture, flavor, processing technique, etc. Based on these, we can categorize Indian spice ingredients under different subheadings, based on the descriptions used in the recipes. For example, coriander is used in recipes very frequently either as a \textbf{fresh herb}, \textbf{dried seeds}, \textbf{powder}, or \textbf{paste}. These are often not interchangeable. Provisions to accommodate all of these are provided in the ontology.
\end{enumerate}

A few examples of the relations supported in the ontology are $\langle \mathbf{r}, \mathbf{has\_ingredient}, \mathbf{i} \rangle$, $\langle \mathbf{r}, \mathbf{has\_cooking\_char}, \mathbf{c} \rangle$ and $\langle \mathbf{r_1}, \mathbf{is\_a}, \mathbf{r} \rangle$ where $\mathbf{r}$, $\mathbf{i}$, $\mathbf{c}$ and $\mathbf{r_1}$ are objects belonging to \textbf{Recipe}, \textbf{Ingredient} and \textbf{Cooking Characteristics} classes and \textbf{Recipe} subclasses respectively. The \textbf{has\_ingredient relation} is a labeled one that stores the measured quantity of ingredient \textbf{i} to be used for recipe \textbf{r} as well.

Different instances of the same recipe may exist with variations in terms of ingredients, instructions, etc. Each of them is considered a unique object of the class recipe. In the Indian context, the same ingredient may be referred to by different regional names. For each such ingredient, a single instance of it is created in the knowledge graph, while storing the different names within it for resolution later. Additionally, since Indian recipes are more compositional in nature in which the \textbf{texture} and \textbf{flavor} of the food item are described more in terms of the end product rather than those of the ingredients alone, we have added texture and flavor as properties of the recipe as well, thus adapting \cite{Haussmann19} to the Indian context. A detailed \textbf{Cuisine} hierarchy has been built for Indian sub-continental food along with associated functions to determine the cuisine label from the recipe ingredients, origin, etc. Variations of the same recipe often exist in different Indian cuisines and it is important to capture the granular cuisines within the Indian subcontinent. For example, \textit{Lucknowi Chicken Biryani} is associated with the \textbf{Awadhi cuisine} of North India whereas \textit{Mutton Donne Biryani} is of South Indian origin and is associated with the \textbf{Karnataka cuisine}. Similarly, mealtime vocabulary has been extended to accommodate Indian festival meals like \textit{Iftaar}, \textit{Navaratri} specials, etc. A long and multidimensional list of \textbf{diet labels} has been also curated from multiple sources to accommodate concepts like <\textbf{Dietary Practice}: \textit{Jain}-vegetarianism>, <\textbf{Health Label}: Keto-friendly>, and <\textbf{Allergen Label}: Dairy-free>. For example, within \textbf{diet labels}, lists of 14 \textbf{dietary practices}, 21 \textbf{allergen labels}, and 22 \textbf{health labels} applicable to the Indian context have been created so far.
\vspace*{-0.1cm}
\begin{enumerate}[noitemsep, resume]
    \item \textbf{Dish} class: Any \textbf{Recipe} instance that is qualified by a measurement unit inherited from the recipe with appropriate scaling and describing the serving size of the recipe is an instance of the \textbf{Dish} class.

    \item \textbf{Platter} class: A platter is a composition of dishes, along with their respective quantities specified, to be always viewed as a single entity.

    \item \textbf{Meal} subclass: Meal is a subclass of platter, which is usually associated with specific occasions or times of day.
    
\end{enumerate}

Now we provide a complete example of these concepts. \textit{Chicken Chettinad} is a popular South Indian recipe of chicken curry. A recipe of \textit{Chicken Chettinad} is mentioned to serve 4 people and contains 1 kg of chicken and various other ingredients like 6 pieces of cardamoms, 2 onions, etc. 1 dish of \textit{Chicken Chettinad} serving 1 person may be a scaled version of the recipe which contains ¼ (or 250 gm of) Chicken. A typical meal called a \textit{South Indian nonvegetarian thali} may be composed of 1 plate of steamed rice, 1 glass of \textit{tomato rasam}, 1 cup of curd, 1 bowl of \textit{Chicken Chettinad}, and 1 \textit{papadum}. Bowl, glass, cup, and plate are typical measures of Indian dishes belonging to different recipe categories which themselves may result in measurement variations. To avoid the ambiguity associated with these terms, we will store a dictionary of these terms along with their precise definitions such as 1 bowl equals 250 gm. A meal can also constitute a single dish alone. For example, \textit{Chicken dum biryani} is a dish that is also a complete meal by itself, prepared using \textit{dum-cook}, a typical Indian cooking process, in a \textit{Handi} which is also a typical Indian cooking vessel.

Mereologically speaking, the \textbf{has\_ingredient} relation between the \textbf{Recipe} and \textbf{Ingredient} classes is the same as FoodOn's \textbf{has ingredient} object property whereas the \textbf{has\_dish} relation between the \textbf{Platter} and \textbf{Dish} classes is the same as FoodOn's \textbf{has part} object property. On the other hand. the \textbf{has\_recipe} relation between the \textbf{Dish} and \textbf{Recipe} classes does not have an equivalent object property in FoodOn and is meant to convey the dynamic scaling of the recipe based on the serving size. Some special relations are also described to capture the essence of a \textbf{Recipe} instance better:
\vspace*{-0.1cm}
\begin{enumerate}[noitemsep]
    \item \textbf{ingr\_is\_substitute\_for} Ingredient property: A pair of ingredients \textbf{i1} and \textbf{i2} are stored as <\textbf{i1}, \textbf{ingr\_is\_substitute\_for}, \textbf{i2}>, if they are substitutes of each other, but may have different \textbf{nutritional properties}, \textbf{diet labels}, etc. For example, Iodized salt and Himalayan pink salt are substitutes, and depending on which one is used, recipe nutrition may vary.

    \item \textbf{recp\_is\_relate\_to} Recipe property: This relation captures the semantic similarity between a pair of recipes. For example, Aloo samosa and Mutton samosa are similar. 
\end{enumerate}

In an ontology, it will be important to specify restrictions, rules, logic, or constraints on concepts and relations that must be satisfied by an object in the ontology for performing consistency checks. The following examples show how restrictions have been used in the current system to validate properties: 
\vspace*{-0.1cm}
\begin{enumerate}[noitemsep]
    \item For all recipes \textbf{r}, if \textbf{r} has the property label “Non-vegetarian” for \textbf{cuisine}, then there exists at least one ingredient \textbf{g} for \textbf{r} that has \textbf{ingredient\_category} "meat" or "egg" from \textbf{animal\_origin}. 
    \item For all recipes \textbf{r}, if \textbf{r} has the property label “Vegetarian” for \textbf{cuisine}, then there does not exist any ingredient \textbf{g} for \textbf{r} that has \textbf{ingredient\_category} "meat" or "egg" from \textbf{animal\_origin}. This defines Indian vegetarian cuisine which is primarily lacto-vegetarian. 
    \item For all recipes r, if \textbf{r} has the property label “Jain” for \textbf{cuisine}, then there does not exist any ingredient \textbf{g} for \textbf{r} that has \textbf{ingredient\_category} ("meat" from \textbf{animal\_origin}) OR ("root\_vegetable" from \textbf{plant\_origin})
\end{enumerate}

In the next section, we define how a knowledge graph of Indian recipes has been built using the proposed ontology design.

\vspace*{-0.2cm}
\section{Knowledge Curation Workflow for building FKG.in}
\vspace*{-0.2cm}
\begin{figure}
  \centering
  \includegraphics[width=0.9\textwidth]{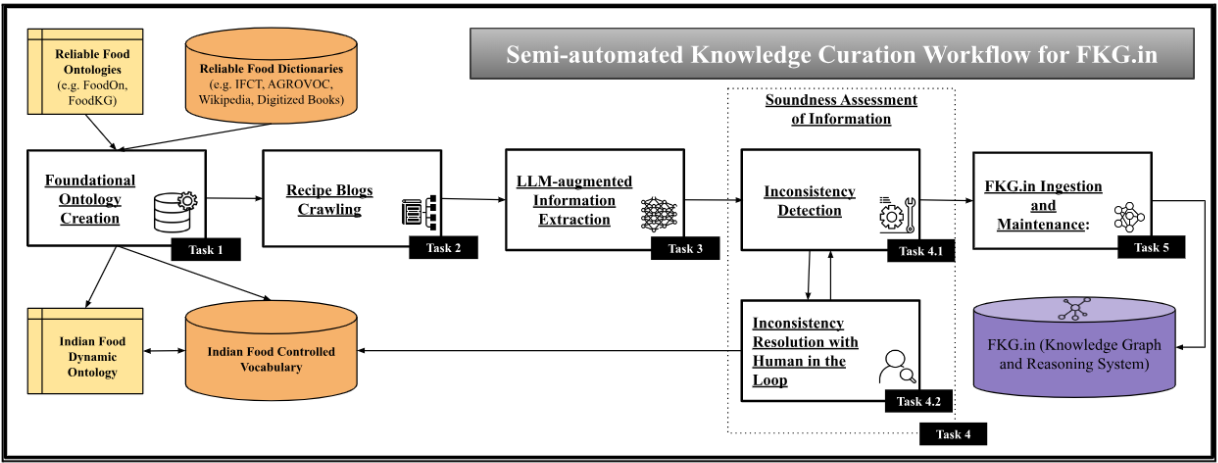}
  \caption{Semi-automated Knowledge Curation Workflow for FKG.in}
    \label{fig:knowledge_curation_workflow}
\end{figure}

\begin{figure}
  \centering
  \includegraphics[width=0.8\textwidth]{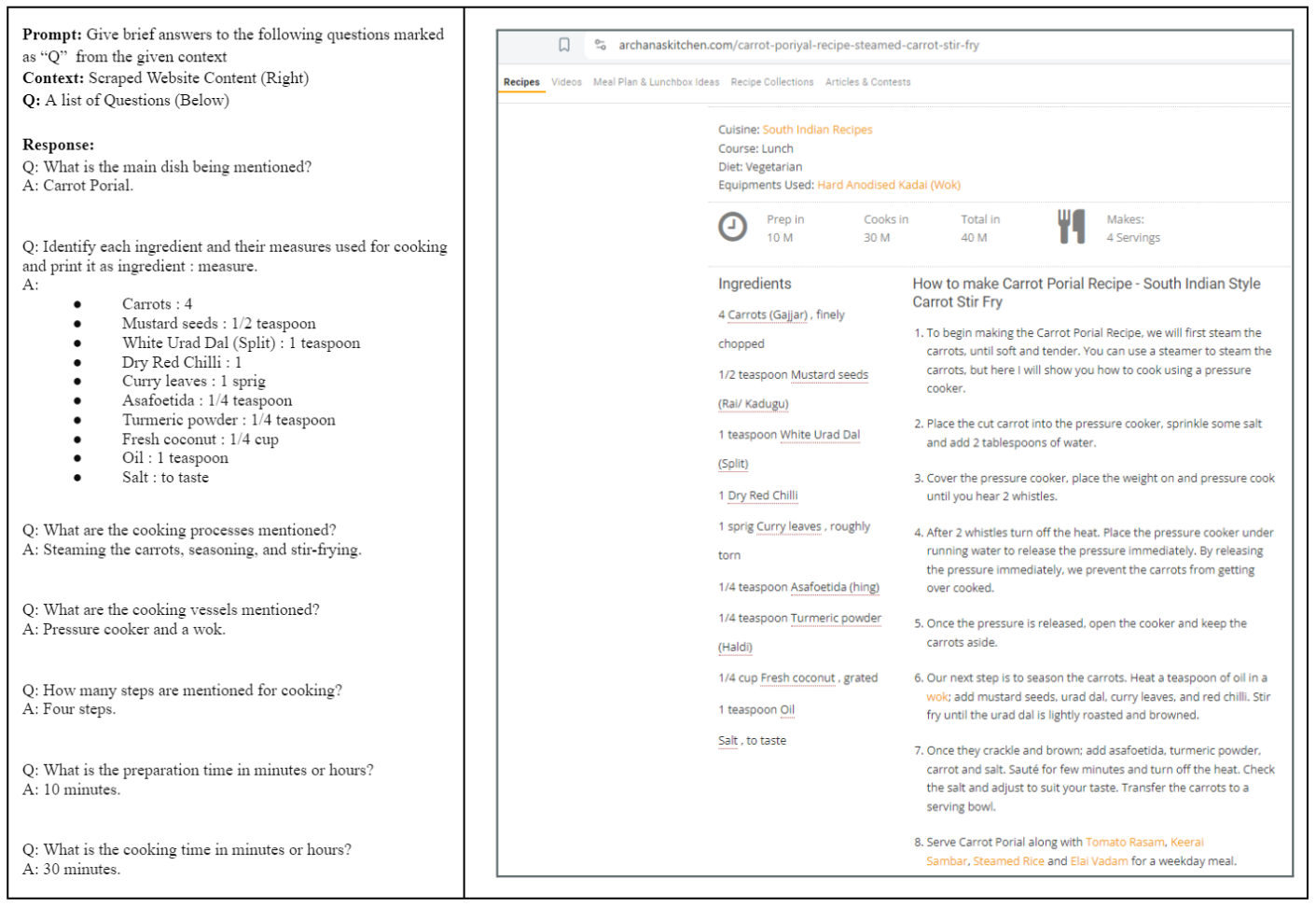}
  \caption{Sample response by ChatGPT based on Contextual Q\&A prompt for an \textit{archanaskitchen} recipe}
  \label{fig:sample_response_by_chatgpt}
\end{figure}

Figure \ref{fig:knowledge_curation_workflow} presents the core tasks of the AI-driven semi-automated workflow for building a food knowledge graph using structured and unstructured information curated from multiple websites. The details of each task are presented in this section.
\vspace*{-0.1cm} 
\begin{itemize}[noitemsep]
    \item \textbf{Task 1 -} Creation of Foundational Ontology
An instance of the foundational ontology for Indian food is created using Web Ontology Language (OWL) in the RDF/XML format by extending reliable and recognized ontologies and dictionaries. Detailed hierarchical designs for storing the labels of \textbf{cuisines}, \textbf{diet labels}, mealtimes, \textbf{cooking characteristics}, etc. described earlier have also been created in the form of a dictionary. Initial vocabulary for each of these structures was populated using various sources like Wikipedia, digitized books \cite{Ashok20}, FoodKG \cite{Haussmann19}, Indian Food Composition Table \cite{Sahu22}, and other sources for Indian food \cite{Padmavathi16}, though the vocabulary itself is not restricted to Indian food alone. The vocabulary was further augmented using large language models. For example, a list of bulb or stem vegetables commonly found in India was generated using OpenAI’s GPT-3.5 Turbo\footnote{https://platform.openai.com/docs/models} and added to the list of ingredients. The initialization step is a one-time process that involves careful manual curation with several sanity checks. The vocabulary is also updated periodically as explained later. 
We are using the SKOS (Simple Knowledge Organization System) W3C recommendation\footnote{https://www.w3.org/2009/08/skos-reference/skos.html} to represent and organize the structured controlled vocabulary as the principal element categories of SKOS such as concepts, labels, notations, documentation, semantic relations, mapping properties, and collections suit the needs of storing food, culinary and nutritional knowledge quite well.

    \item \textbf{Task 2 -} Crawling of Recipe Blogs:
We have identified 40 recipe blogs and websites with rich information about Indian food recipes, their nutritional information, and other culinary information. To begin with, we have crawled 5 recipe blogs viz. \textbf{archanaskitchen}, \textbf{hookedonheat}, \textbf{indianhealthyrecipes}, \textbf{masalakorb} and \textbf{vegrecipesofindia}, each of which has several recipe websites along with a detailed recipe card for each. The crawler gathers content from each recipe blog page and stores it locally as an HTML file along with metadata like its \textbf{source URL}, \textbf{recipe name}, \textbf{recipe category}, \textbf{blogpost timestamp}, and \textbf{scraping timestamp} for reproducibility and parsing.

    \item \textbf{Task 3 -} LLM-augmented Information Extraction:
The HTML files are then cleaned and parsed to extract the recipe details such as ingredients, cooking characteristics, nutritional information, etc. from both structured and unstructured parts. This was done by setting up a pipeline using Langchain and GPT-3.5 Turbo, a large language model, to process the recipe webpage content and generate semi-structured output using zero-shot and few-shot prompts. GPT 3.5 is also employed to translate Indian ingredient names, written in Indian or Roman scripts to their English names. This helps in entity resolution and consolidation later while populating the knowledge graph after the soundness check. This process is executed for all the recipe URLs before moving on to the next steps. 

    \item \textbf{Task 4 -} Soundness Assessment of Information:
After generating food entities and relations from each recipe, this step runs automated checks to validate the information against existing vocabularies, performs entity resolution and then flags possible inconsistencies to humans for correction. This step is to ensure that the information added to the knowledge base is correct.
\vspace*{-0.1cm}
    \begin{itemize}[noitemsep]
        \item \textbf{Task 4.1 -} Figure \ref{fig:sample_response_by_chatgpt} shows a sample output. The extracted entities as part of ingredient, cooking processes, cooking utensils etc. are clustered using Locality Sensitive Hashing (LSH) for entity resolution to improve the accuracy and precision of entity lists. The consolidated lists are cross-checked against existing known entries. Unknown or new information are verified through human intervention. For example, if an entity \textit{kadahi (wok)} is incorrectly identified as a recipe ingredient, instead of a vessel to cook as listed in the vocabulary, then the system flags an inconsistency. Similarly, Indian spice names, if already present in the vocabulary, are mapped to the unique identifiers and if not present, then flagged for human inspection to be included in the vocabulary appropriately. Restriction-based checks are also applied at this stage. 

        \item \textbf{Task 4.2 -} All inconsistencies identified in the earlier step are presented to human curators for validation and correction if needed. For example, a common mistake made by language tools, including LLMs, is the failure to detect multi-word named entities correctly. For example, for a recipe to make "\textit{pudina chutney sandwich}", the key ingredient is "\textit{pudina chutney}" and not “\textit{pudina}” which is a basic ingredient. A human can correct the entity and help in appropriate incorporation of “\textit{Pudina chutney}” as an ingredient, which is also a recipe, and may appear as such in other recipes also. Several instances of incorrect and incomplete information extraction are observed for unstructured portions. 

        An easy-to-use interface has been built to aid the correction process. As expected, the number of entities that need human intervention, go down. Further, insights obtained from human intervention were used to improve the LLM prompts, which also helped reduce the error of the extraction process. Human feedback is also used to augment the ontology in an atomic, reliable, and consistent manner to accommodate new information obtained from the recipe web pages.
    \end{itemize}

The above methods ensure the soundness of the knowledge graph, i.e. information added to the knowledge base is correct. It however does not ensure completeness of information in situations where the large language model fails to extract a piece of information altogether. Such issues will be addressed in the future while working on the completeness of the knowledge graph for Indian food.
    
    \item \textbf{Task 5 -} FKG.in Ingestion and Maintenance:
The verified and validated information components are ingested into the knowledge graph. While some of them may result in vocabulary extensions, some are added as instances of classes and relationships.
\end{itemize}

The algorithmic details of the knowledge curation workflow are presented below: 
\vspace*{-0.1cm}
\begin{enumerate}[noitemsep]
    \item \textbf{Initialization:} Make a list of target information to be extracted from recipe URLs based on the data/class properties associated with ingredients and recipes as per the ontology.

    \item \textbf{Crawling and Extraction:} Fetch and store the recipe dump locally for all the recipe URLs. Use the \textbf{requests} library in Python to parse and extract target information from the recipe card and store it in an XML file.

    \item \textbf{Semantic Resolution:} Use semantic resolution to map property names across recipe domains. For example, recipe blogs may use the term \textbf{region} or \textbf{style} to refer to \textbf{cuisine}. While the dataset is curated mostly with manual intervention in the initial phases, the lists of property names and values are automated and learned over time.

    \item \textbf{LLM-enabled Entity Recognition:} Use fine-tuned prompts and LLMs to extract information and recognize entities from the unstructured recipe webpage content which contains ingredient details along with cooking instructions, cooking characteristics, etc in long text. Ingredient measures are also included in this information. Prompt engineering is used to store the extracted information in a structured format to enable comparison with the recipe card information that was stored in the XML format earlier.

    \item \textbf{Soundness Assessment:} Compare the XML output with the LLM output to obtain a match score, where a score of +1 indicates a match between the two tuples and a score of -1, whenever a mismatch occurs between a recipe card tuple and an LLM output. All LLM tuples that do not have a corresponding match in the recipe card are matched against the vocabulary terms of the corresponding property list. For each match found, a score of +1 is awarded and -1 for matches not found. For each recipe parsed, the total positive score is an indicator of the soundness of the information as it is double-checked against recipe cards and vocabulary lists. All negative scores are flagged for human validation in the next step. All XML tuples and LLM tuples with a score of +1 are candidate elements for the knowledge graph. The total positive score provides an assessment of the underlying LLM-based information extraction system, which will be the only way to extract information from totally unstructured websites without recipe cards. This will be explored in future.

    \item \textbf{Human Validation and Updation:} An easy-to-use interface is used by human curator to assess all flagged information. Vocabulary updates, if any, are also enabled through an interactive platform. Tuples are also assessed and corrected, if necessary. All actions, resolved and not resolved, are documented. The information was used heavily to finalize the ontology design and is stored for any future needs.

    \item \textbf{Ingestion of Tuples into FKG.in:} After updating the vocabularies, the knowledge graph is ingested with the new and validated information as per the latest ontology. All unique tuples from the candidate set of step 5 and human-approved tuples from step 6 are used to extend the knowledge graph to include new instances of objects and relations. The tuples in the form of RDF/XML triple-stores are stored in an OWL file which stores both the Indian food ontology and the associated vocabulary. We are currently using Ontotext’s GraphDB\footnote{https://graphdb.ontotext.com/} to build the knowledge graph.
\end{enumerate}

Though not implemented currently, in the future the knowledge graph will undergo systematic checks to perform evaluations and optimizations by using quality metrics and optimized organization principles based on the SKOS recommendation.

\vspace*{-0.2cm}
\section{Current Status of FKG.in}
\vspace*{-0.2cm}
The size of FKG.in is presently around 50 MB. It has information about 9628 unique recipe instances gathered from the five recipe blogs mentioned earlier. After consolidating the metadata, these recipes belong to 39 distinct categories such as \textbf{breakfast}, \textbf{cakes}, \textbf{vegetarian}, \textbf{Hyderabadi}, \textbf{Indian sweets}, etc. The total number of ingredient nodes in the knowledge graph is currently 38819. We have observed some nodes have Hindi names in \textbf{Devanagari} fonts, indicating that more resolution rules will need to be added to address code-mixing. It has also been observed that complex ingredients, which are recipes themselves, are duplicated in the knowledge graph, as both a recipe node and an ingredient node. This needs to be resolved with an associative relationship, which is currently not a part of the design.

\vspace*{-0.2cm}
\section{Conclusions and Future Work}
\vspace*{-0.2cm}
In this paper, we have presented the work initiated towards building FKG.in. Due to the lack of reference resources, almost everything had to be initiated from scratch. Unlike earlier methods, which have focused on building knowledge graphs from semi-structured data and in application-specific ways, our focus is on using AI-enabled methods for extracting relevant information from all kinds of recipe blogs to populate the knowledge graph. Zero-shot and few-shot methods that exploit the large language model GPT-3.5 Turbo have been used extensively to build initial vocabularies and subsequently to extract entities and relations to populate the knowledge base. Methods to ensure soundness of information are also incorporated into the pipeline. We have presented the current status of the knowledge graph called FKG.in.

Further work on the refinement of ontology design as well as on knowledge engineering techniques is underway. NLP tools for multilingual semantic reasoning are one of the primary areas identified for future research. Another area of focus is on quantitative assessments of the soundness and completeness of the knowledge graph. 

In the future, we expect work to spread across many other directions that involve reasoning over Indian food concepts as well. Such knowledge graphs can address several questions of historical, social, and cultural aspects of food and food habits, enable several applications including but not limited to food recommendation systems, personal health navigation systems, recipe generation, and recipe recommendation systems, and also aid knowledge discovery from underlying data. The methods for knowledge curation proposed in this paper are generic and can be replicated for any domain.

\vspace*{-0.2cm}
\section{Acknowledgments}
\vspace*{-0.2cm}



This research was supported by the Ashoka Mphasis Lab - a collaboration between Ashoka University and Mphasis Limited.






\bibliography{fkg_in}

\begin{thebibliography}{25}
\expandafter\ifx\csname natexlab\endcsname\relax\def\natexlab#1{#1}\fi
\providecommand{\url}[1]{\texttt{#1}}
\providecommand{\href}[2]{#2}
\providecommand{\path}[1]{#1}
\providecommand{\DOIprefix}{doi:}
\providecommand{\ArXivprefix}{arXiv:}
\providecommand{\URLprefix}{URL: }
\providecommand{\Pubmedprefix}{pmid:}
\providecommand{\doi}[1]{\href{http://dx.doi.org/#1}{\path{#1}}}
\providecommand{\Pubmed}[1]{\href{pmid:#1}{\path{#1}}}
\providecommand{\bibinfo}[2]{#2}
\ifx\xfnm\relax \def\xfnm[#1]{\unskip,\space#1}\fi
\bibitem[{Min et~al.(2019)Min, Jiang, Liu, Rui, and Jain}]{Min19}
\bibinfo{author}{W.~Min}, \bibinfo{author}{S.~Jiang}, \bibinfo{author}{L.~Liu}, \bibinfo{author}{Y.~Rui}, \bibinfo{author}{R.~Jain},
\newblock \bibinfo{title}{A survey on food computing},
\newblock \bibinfo{journal}{{ACM} Comput. Surveys} \bibinfo{volume}{52} (\bibinfo{year}{2019}) \bibinfo{pages}{1--36}. \DOIprefix\doi{10.1145/3329168}.
\bibitem[{Min et~al.(2020)Min, Jiang, and Jain}]{Min20}
\bibinfo{author}{W.~Min}, \bibinfo{author}{S.~Jiang}, \bibinfo{author}{R.~Jain},
\newblock \bibinfo{title}{Food recommendation: Framework, existing solutions, and challenges},
\newblock \bibinfo{journal}{{IEEE} Transactions on Multimedia} \bibinfo{volume}{22} (\bibinfo{year}{2020}) \bibinfo{pages}{2659--2671}. \DOIprefix\doi{10.1109/TMM.2019.2958761}.
\bibitem[{Min et~al.(2022)Min, Liu, Xu, and Jiang}]{Min22}
\bibinfo{author}{W.~Min}, \bibinfo{author}{C.~Liu}, \bibinfo{author}{L.~Xu}, \bibinfo{author}{S.~Jiang},
\newblock \bibinfo{title}{Applications of knowledge graphs for food science and industry},
\newblock \bibinfo{journal}{Patterns} \bibinfo{volume}{3} (\bibinfo{year}{2022}). \DOIprefix\doi{10.1016/j.patter.2022.100484}.
\bibitem[{Cordier et~al.(2014)Cordier, Dufour-Lussier, Lieber, Nauer, Badra, Cojan, Gaillard, Infante-Blanco, Molli, Napoli, and Skaf-Molli}]{Cordier14}
\bibinfo{author}{A.~Cordier}, \bibinfo{author}{V.~Dufour-Lussier}, \bibinfo{author}{J.~Lieber}, \bibinfo{author}{E.~Nauer}, \bibinfo{author}{F.~Badra}, \bibinfo{author}{J.~Cojan}, \bibinfo{author}{E.~Gaillard}, \bibinfo{author}{L.~Infante-Blanco}, \bibinfo{author}{P.~Molli}, \bibinfo{author}{A.~Napoli}, \bibinfo{author}{H.~Skaf-Molli},
\newblock \bibinfo{title}{Taaable: a case-based system for personalized cooking},
\newblock in: \bibinfo{editor}{S.~Montani}, \bibinfo{editor}{L.~C. Jain} (Eds.), \bibinfo{booktitle}{Successful Case-based Reasoning Applications-2}, \bibinfo{publisher}{Springer Berlin Heidelberg}, \bibinfo{address}{Berlin, Heidelberg}, \bibinfo{year}{2014}, pp. \bibinfo{pages}{121--162}. \DOIprefix\doi{10.1007/978-3-642-38736-4_7}.
\bibitem[{Ribeiro et~al.(2006)Ribeiro, Batista, Pardal, Mamede, and Pinto}]{Ribeiro06}
\bibinfo{author}{R.~Ribeiro}, \bibinfo{author}{F.~Batista}, \bibinfo{author}{J.~P. Pardal}, \bibinfo{author}{N.~J. Mamede}, \bibinfo{author}{H.~S. Pinto},
\newblock \bibinfo{title}{Cooking an ontology},
\newblock in: \bibinfo{editor}{J.~Euzenat}, \bibinfo{editor}{J.~Domingue} (Eds.), \bibinfo{booktitle}{Artificial Intelligence: Methodology, Systems, and Applications}, \bibinfo{publisher}{Springer}, \bibinfo{address}{Berlin}, \bibinfo{year}{2006}, pp. \bibinfo{pages}{213--221}. \DOIprefix\doi{10.1007/11861461_23}.
\bibitem[{Sherimon et~al.(2021)Sherimon, P.C., Ismaeel, Varkey, and B.}]{Sherimon21}
\bibinfo{author}{V.~Sherimon}, \bibinfo{author}{S.~P.C.}, \bibinfo{author}{A.~Ismaeel}, \bibinfo{author}{W.~Varkey}, \bibinfo{author}{N.~B.},
\newblock \bibinfo{title}{Modeling of seafood domain using ontology},
\newblock \bibinfo{journal}{Intl Jr. of Open Info. Technologies} \bibinfo{volume}{9} (\bibinfo{year}{2021}).
\bibitem[{Zulaika et~al.(2018)Zulaika, Gutiérrez, and López-de Ipiña}]{Zulaika18}
\bibinfo{author}{U.~Zulaika}, \bibinfo{author}{A.~Gutiérrez}, \bibinfo{author}{D.~López-de Ipiña},
\newblock \bibinfo{title}{Enhancing profile and context aware relevant food search through knowledge graphs},
\newblock \bibinfo{journal}{Proceedings} \bibinfo{volume}{2} (\bibinfo{year}{2018}). \DOIprefix\doi{10.3390/proceedings2191228}.
\bibitem[{Lei et~al.(2021)Lei, Haq, Zeb, Suzauddola, and Zhang}]{Lei21}
\bibinfo{author}{Z.~Lei}, \bibinfo{author}{A.~U. Haq}, \bibinfo{author}{A.~Zeb}, \bibinfo{author}{M.~Suzauddola}, \bibinfo{author}{D.~Zhang},
\newblock \bibinfo{title}{Is the suggested food your desired?: Multi-modal recipe recommendation with demand-based knowledge graph},
\newblock \bibinfo{journal}{Expert Systems with Applications} \bibinfo{volume}{186} (\bibinfo{year}{2021}). \DOIprefix\doi{10.1016/j.eswa.2021.115708}.
\bibitem[{Marín et~al.(2021)Marín, Biswas, Ofli, Hynes, Salvador, Aytar, Weber, and Torralba}]{Marin21}
\bibinfo{author}{J.~Marín}, \bibinfo{author}{A.~Biswas}, \bibinfo{author}{F.~Ofli}, \bibinfo{author}{N.~Hynes}, \bibinfo{author}{A.~Salvador}, \bibinfo{author}{Y.~Aytar}, \bibinfo{author}{I.~Weber}, \bibinfo{author}{A.~Torralba},
\newblock \bibinfo{title}{Recipe1m+: A dataset for learning cross-modal embeddings for cooking recipes and food images},
\newblock \bibinfo{journal}{IEEE Trans. on PAMI} \bibinfo{volume}{43} (\bibinfo{year}{2021}) \bibinfo{pages}{187--203}. \DOIprefix\doi{10.1109/TPAMI.2019.2927476}.
\bibitem[{Bagler(2020)}]{Bagler20}
\bibinfo{author}{G.~Bagler},
\newblock \bibinfo{title}{A generative grammar of cooking},
\newblock \bibinfo{journal}{arXiv}  (\bibinfo{year}{2020}). \DOIprefix\doi{10.48550/arXiv.2211.09059}.
\bibitem[{Batra et~al.(2020)Batra, Diwan, Upadhyay, Kalra, Sharma, Sharma, Khanna, Marwah, Kalathil, Singh, Tuwani, and Bagler}]{Batra20}
\bibinfo{author}{D.~Batra}, \bibinfo{author}{N.~Diwan}, \bibinfo{author}{U.~Upadhyay}, \bibinfo{author}{J.~S. Kalra}, \bibinfo{author}{T.~Sharma}, \bibinfo{author}{A.~K. Sharma}, \bibinfo{author}{D.~Khanna}, \bibinfo{author}{J.~S. Marwah}, \bibinfo{author}{S.~Kalathil}, \bibinfo{author}{N.~Singh}, \bibinfo{author}{R.~Tuwani}, \bibinfo{author}{G.~Bagler},
\newblock \bibinfo{title}{{RecipeDB}: a resource for exploring recipes},
\newblock \bibinfo{journal}{Database} \bibinfo{volume}{2020} (\bibinfo{year}{2020}). \DOIprefix\doi{10.1093/database/baaa077}.
\bibitem[{Cantais et~al.(2005)Cantais, Dominguez, Gigante, Laera, and Tamma}]{Cantais05}
\bibinfo{author}{J.~Cantais}, \bibinfo{author}{D.~Dominguez}, \bibinfo{author}{V.~Gigante}, \bibinfo{author}{L.~Laera}, \bibinfo{author}{V.~Tamma},
\newblock \bibinfo{title}{An example of food ontology for diabetes control},
\newblock \bibinfo{year}{2005}, pp. \bibinfo{pages}{1--9}.
\bibitem[{Snae and Bruckner(2008)}]{Snae08}
\bibinfo{author}{C.~Snae}, \bibinfo{author}{M.~Bruckner},
\newblock \bibinfo{title}{Foods: A food-oriented ontology-driven system},
\newblock in: \bibinfo{booktitle}{2008 2nd IEEE Int'l Conf. on Digital Ecosystems \& Technologies}, \bibinfo{year}{2008}, pp. \bibinfo{pages}{168--176}. \DOIprefix\doi{10.1109/DEST.2008.4635195}.
\bibitem[{Dooley et~al.(2018)Dooley, Griffiths, Gosal, Buttigieg, Hoehndorf, Lange, Schriml, Brinkman, and Hsiao}]{Dooley18}
\bibinfo{author}{D.~M. Dooley}, \bibinfo{author}{E.~J. Griffiths}, \bibinfo{author}{G.~S. Gosal}, \bibinfo{author}{P.~L. Buttigieg}, \bibinfo{author}{R.~Hoehndorf}, \bibinfo{author}{M.~C. Lange}, \bibinfo{author}{L.~M. Schriml}, \bibinfo{author}{F.~S.~L. Brinkman}, \bibinfo{author}{W.~W.~L. Hsiao},
\newblock \bibinfo{title}{{FoodOn}: a harmonized food ontology to increase global food traceability, quality control and data integration},
\newblock \bibinfo{journal}{npj Science of Food} \bibinfo{volume}{2} (\bibinfo{year}{2018}). \DOIprefix\doi{10.1038/s41538-018-0032-6}.
\bibitem[{Haussmann et~al.(2019)Haussmann, Seneviratne, Chen, Ne'eman, Codella, Chen, McGuinness, and Zaki}]{Haussmann19}
\bibinfo{author}{S.~Haussmann}, \bibinfo{author}{O.~Seneviratne}, \bibinfo{author}{Y.~Chen}, \bibinfo{author}{Y.~Ne'eman}, \bibinfo{author}{J.~Codella}, \bibinfo{author}{C.-H. Chen}, \bibinfo{author}{D.~L. McGuinness}, \bibinfo{author}{M.~J. Zaki},
\newblock \bibinfo{title}{{FoodKG}: A semantics-driven knowledge graph for food recommendation},
\newblock in: \bibinfo{editor}{C.~Ghidini}, \bibinfo{editor}{O.~Hartig}, \bibinfo{editor}{M.~Maleshkova}, \bibinfo{editor}{V.~Sv{\'a}tek}, \bibinfo{editor}{I.~Cruz}, \bibinfo{editor}{A.~Hogan}, \bibinfo{editor}{J.~Song}, \bibinfo{editor}{M.~Lefran{\c{c}}ois}, \bibinfo{editor}{F.~Gandon} (Eds.), \bibinfo{booktitle}{The Semantic Web -- ISWC 2019}, \bibinfo{publisher}{Springer International Publishing}, \bibinfo{address}{Cham}, \bibinfo{year}{2019}, pp. \bibinfo{pages}{146--162}. \DOIprefix\doi{10.1007/978-3-030-30796-7_10}, \bibinfo{note}{resource Website: https://foodkg.github.io}.
\bibitem[{Chi et~al.(2018)Chi, Yu, Qi, and Xu}]{Chi18}
\bibinfo{author}{Y.~Chi}, \bibinfo{author}{C.~Yu}, \bibinfo{author}{X.~Qi}, \bibinfo{author}{H.~Xu},
\newblock \bibinfo{title}{Knowledge management in healthcare sustainability: A smart healthy diet assistant in traditional chinese medicine culture},
\newblock \bibinfo{journal}{Sustainability} \bibinfo{volume}{10} (\bibinfo{year}{2018}). \DOIprefix\doi{10.3390/su10114197}.
\bibitem[{Caracciolo et~al.(2011)Caracciolo, Stellato, Rajbahndari, Morshed, Johannsen, Jaques, and Keizer}]{Caracciolo11}
\bibinfo{author}{C.~Caracciolo}, \bibinfo{author}{A.~Stellato}, \bibinfo{author}{S.~Rajbahndari}, \bibinfo{author}{A.~Morshed}, \bibinfo{author}{G.~Johannsen}, \bibinfo{author}{Y.~Jaques}, \bibinfo{author}{J.~Keizer},
\newblock \bibinfo{title}{Thesaurus maintenance, alignment and publication as linked data: The agroovoc use case},
\newblock in: \bibinfo{editor}{E.~Garc{\'i}a-Barriocanal}, \bibinfo{editor}{Z.~Cebeci}, \bibinfo{editor}{M.~C. Okur}, \bibinfo{editor}{A.~{\"O}zt{\"u}rk} (Eds.), \bibinfo{booktitle}{Metadata and Semantic Research}, \bibinfo{publisher}{Springer Berlin Heidelberg}, \bibinfo{address}{Berlin, Heidelberg}, \bibinfo{year}{2011}, pp. \bibinfo{pages}{489--499}. \DOIprefix\doi{10.1007/978-3-642-24731-6_48}.
\bibitem[{Pizzuti et~al.(2014)Pizzuti, Mirabelli, Sanz-Bobi, and Goméz-Gonzaléz}]{Pizzuti14}
\bibinfo{author}{T.~Pizzuti}, \bibinfo{author}{G.~Mirabelli}, \bibinfo{author}{M.~A. Sanz-Bobi}, \bibinfo{author}{F.~Goméz-Gonzaléz},
\newblock \bibinfo{title}{Food track \& trace ontology for helping the food traceability control},
\newblock \bibinfo{journal}{Journal of Food Engineering} \bibinfo{volume}{120} (\bibinfo{year}{2014}) \bibinfo{pages}{17--30}. \DOIprefix\doi{10.1016/j.jfoodeng.2013.07.017}.
\bibitem[{Ameri et~al.(2022)Ameri, Wallace, Yoder, and Riddick}]{Ameri22}
\bibinfo{author}{F.~Ameri}, \bibinfo{author}{E.~Wallace}, \bibinfo{author}{R.~Yoder}, \bibinfo{author}{F.~Riddick},
\newblock \bibinfo{title}{Enabling traceability in agri-food supply chains using an ontological approach},
\newblock \bibinfo{journal}{Jr. of Computing and Info. Sc. in Engg.} \bibinfo{volume}{22} (\bibinfo{year}{2022}) \bibinfo{pages}{17--30}. \DOIprefix\doi{10.1115/1.4054092}.
\bibitem[{Qin et~al.(2019)Qin, Hao, and Zhao}]{Qin19}
\bibinfo{author}{L.~Qin}, \bibinfo{author}{Z.~Hao}, \bibinfo{author}{L.~Zhao},
\newblock \bibinfo{title}{Food safety knowledge graph and question answering system},
\newblock in: \bibinfo{booktitle}{ICIT '19: Proc. of the 2019 7th Intl Conf. on Info. Tech.: IoT and Smart City}, \bibinfo{publisher}{ACM}, \bibinfo{address}{New York, United States}, \bibinfo{year}{2019}, pp. \bibinfo{pages}{559--–564}. \DOIprefix\doi{10.1145/3377170.3377260}.
\bibitem[{Qin et~al.(2020)Qin, Hao, and Zhao}]{Qin20}
\bibinfo{author}{L.~Qin}, \bibinfo{author}{Z.~Hao}, \bibinfo{author}{L.~Zhao},
\newblock \bibinfo{title}{Question answering system based on food spot-check knowledge graph},
\newblock in: \bibinfo{booktitle}{ICCDE '20: Proceedings of 2020 6th International Conference on Computing and Data Engineering}, \bibinfo{publisher}{Association for Computing Machinery}, \bibinfo{address}{New York, United States}, \bibinfo{year}{2020}, pp. \bibinfo{pages}{168–--172}. \DOIprefix\doi{10.1145/3379247.3379292}.
\bibitem[{Padmavathi and Krishnamurthy(2016)}]{Padmavathi16}
\bibinfo{author}{T.~Padmavathi}, \bibinfo{author}{M.~Krishnamurthy},
\newblock \bibinfo{title}{Ontology for the domain of food science},
\newblock \bibinfo{journal}{Journal of Information and Knowledge} \bibinfo{volume}{53} (\bibinfo{year}{2016}) \bibinfo{pages}{409–--417}. \DOIprefix\doi{10.17821/srels/2016/v53i5/89230}.
\bibitem[{Sahu(2022)}]{Sahu22}
\bibinfo{author}{S.~Sahu}, \bibinfo{title}{ifct2017/compositions: Detailed nutrient composition of 528 key foods in india}, \bibinfo{year}{2022}.
\bibitem[{Madalli et~al.(2017)Madalli, Chatterjee, and Dutta}]{Madalli17}
\bibinfo{author}{D.~P. Madalli}, \bibinfo{author}{U.~Chatterjee}, \bibinfo{author}{B.~Dutta},
\newblock \bibinfo{title}{An analytical approach to building a core ontology for food},
\newblock \bibinfo{journal}{Journal of Documentation} \bibinfo{volume}{70} (\bibinfo{year}{2017}) \bibinfo{pages}{123--144}. \DOIprefix\doi{10.1108/JD-02-2016-0015}.
\bibitem[{Ashok(2020)}]{Ashok20}
\bibinfo{author}{K.~Ashok}, \bibinfo{title}{Masala Lab: The Science of Indian Cooking}, \bibinfo{edition}{{Kindle}} ed., \bibinfo{publisher}{Penguin}, \bibinfo{year}{2020}.

\end{thebibliography}


\appendix

\end{document}